%% file: 00.main.tex
\documentclass[
]{ceurart}

\usepackage{listings}
\usepackage{graphicx}
\usepackage{caption}
\usepackage[most]{tcolorbox}
\usepackage{placeins}
\usepackage{kotex} 

\newcommand{\algname}{MPR-CiteG}

\begin{document}

\copyrightyear{2025}
\copyrightclause{Copyright for this paper by its authors.
  Use permitted under Creative Commons License Attribution 4.0
  International (CC BY 4.0).}

\conference{The 1st International Workshop on Retrieval-Driven Generative AI \& ScienceON AI Challenge 2025}

\title{MPR-CiteG: Enhancing RAG with Multi-Portfolio Retrieval and Citation-Grounded Generation}

\tnotemark[1]

\author[1]{Hyewon Lee}[%
email=noweyh927@g.cnu.ac.kr,
]
\fnmark[1]
\address[1]{Data Intelligence Laboratory, Department of Computer Science and Engineering, Chungnam National University, Daejeon 34134, Republic of Korea}


\author[1]{Minkyung Song}[%
email=kyung@o.cnu.ac.kr,
]
\fnmark[1]

\author[1]{Junghyun Oh}[%
email=ojh7839@o.cnu.ac.kr,
]

\author[1]{Seunghoon Han}[%
email=tmdgns129@g.cnu.ac.kr,
]

\author[1]{Sungsu Lim}[%
email=sungsu@cnu.ac.kr,
]
\cormark[1]
\cortext[1]{Corresponding author.}
\fntext[1]{These authors contributed equally.}

\begin{abstract}
  This paper presents the \algname{} framework, which achieved second place in the ScienceON AI Challenge by addressing two fundamental challenges in generative AI: inefficient retrieval and the absence of source verification. We propose a dual-component system, termed \algname{}, in which the Multi-Portfolio Retriever (MPR) efficiently retrieves diverse and relevant information, while the Citation-Grounded Generation (CiteG) module ensures that every generated output remains factually consistent and explicitly attributed to its source. \algname{} represents a significant step toward building more trustworthy and accurate LLMs that are not only capable of generating information but also of grounding their responses in reliable evidence, thereby mitigating common issues like model hallucination. Extensive experiments on the challenge dataset validate the effectiveness and reliability of our approach. Our code is available at \href{https://github.com/2noweyh/MPR-citeG}{https://github.com/2noweyh/MPR-citeG}.
\end{abstract}


\maketitle

\input{01.int}

\input{03.met} 
\input{04.exp}

\input{05.con} 

\begin{acknowledgments}
  This work was supported by Institute of Information \& communications Technology Planning \& Evaluation (IITP) grant funded by the Korea government(MSIT) (No.RS-2022-00155857, Artificial Intelligence Convergence Innovation Human Resources Development (Chungnam National University)) 
\end{acknowledgments}

\bibliography{reference}

\appendix
\input{06.app}
\clearpage

\end{document}

%% file: 01.int.tex
\section{Introduction}
Recent advances in Large Language Models (LLMs)~\cite{zhao2023survey} have achieved success in retrieval-augmented tasks such as summarization and question answering. However, when relying solely on their internal knowledge, LLMs face inherent limitations in providing up-to-date and factually grounded responses~\cite{huang2025survey}. To address this issue, Retrieval-Augmented Generation (RAG)~\cite{gao2023retrieval} enables models to leverage retrieved results as evidence to produce accurate and trustworthy outputs. In this context, the ScienceON AI Challenge~\cite{sai-challenge} benchmarks RAG systems on generating responses from public search results returned by the ScienceON OpenAPI, requiring participants to retrieve documents, re-rank results, and produce evidence-grounded answers under a fixed GPU budget—making efficiency crucial.

Prior studies improve RAG along two axes—query refinement and citation faithfulness—but important gaps persist. Query-focused methods~\cite{chan2024rq, li2024dmqr} enhance retrieval recall through rewriting and diversification, yet overlook practical efficiency concerns such as budgeting retrieval cost and coordinating heterogeneous retrievers under limited compute. Meanwhile, citation-oriented approaches~\cite{qian2025vericite, cai2024forag} improve faithfulness but rely on resource-intensive, multi-stage pipelines. These methods, which use reinforcement learning from human feedback or natural language inference-based verification, remain ill-suited to public-API environments where snippet-level grounding and latency constraints dominate.

We address these gaps with \algname{}, which couples a Multi-Portfolio Retriever (MPR) and Citation-Grounded Generation (CiteG) to convert constraints into performance gains. MPR overcomes budget/coordinator limits by expanding each query into four parallel Query Portfolios (\textbf{\textit{Q}}, \textbf{\textit{S}}, \textbf{\textit{F}}, and \textbf{\textit{G}}), then de-duplicating and re-ranking the merged pool with a cross-encoder~\cite{nogueira2019passage} to yield a compact, high-precision context. CiteG overcomes citation/latency costs with a two-stage pipeline: first, it generates a document-conditioned draft from top evidence; then it aligns sentences to sources and assigns markers, producing fluent answers with citations. Together, these components deliver accurate, citation-backed responses within the fixed-GPU regime and the public ScienceON OpenAPI.

%% file: 03.met.tex
\section{Architecture}

\begin{figure}[t]
    \centering
    \includegraphics[width=1.0\columnwidth]{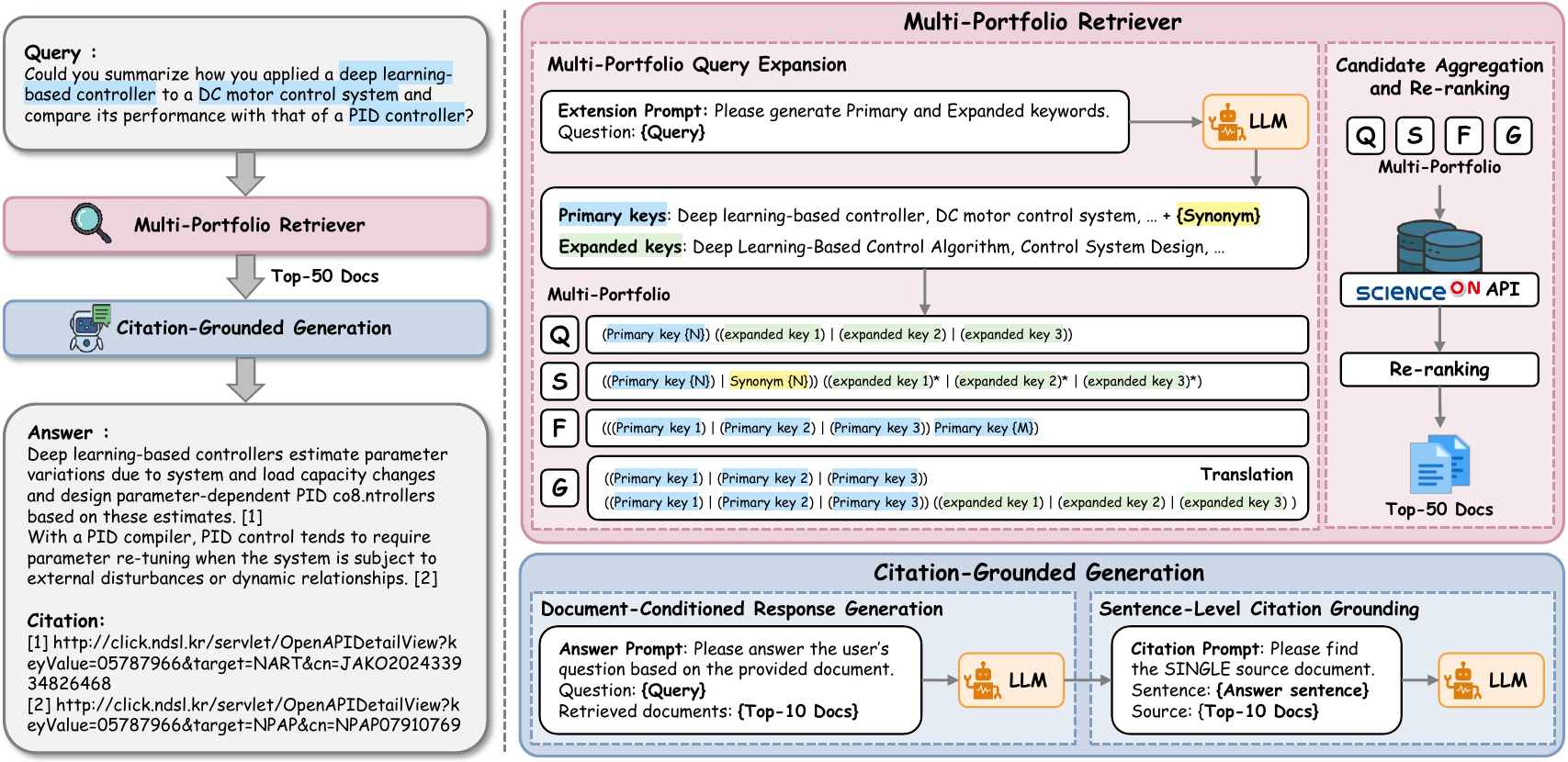}
    \caption{Overall framework consisting of Multi-Portfolio Retriever and Citation-Grounded Generation.}
    \label{fig:framework}
\end{figure}

In this section, we present the overall architecture that combines the Multi-Portfolio Retriever (MPR) with the Citation-Grounded Generation (CiteG) to achieve both efficient retrieval and source-grounded generation. As illustrated in Fig.~\ref{fig:framework}, the top-right block depicts MPR expanding the initial query into multiple portfolios, after which the retrieved documents are aggregated and re-ranked. The bottom-right block shows CiteG generating the final user-facing answer from the retrieved documents with explicit citations, ensuring factual accuracy and proper referencing throughout the scientific context.

\subsection{Multi-Portfolio Retriever}
Our RAG retriever follows a multi-stage design that maximizes recall in the early phase and ensures precision in the final phase. We call this the Multi-Portfolio Retriever (MPR), which systematically expands, diversifies, and refines queries to collect high-quality contextual information for reliable answer generation. The process is organized into two main components: Section~\ref{section:Multi-Portfolio} Multi-Portfolio Query Expansion, and Section~\ref{section:ReRanking} Candidate Aggregation and Re-Ranking.

\subsubsection{Multi-Portfolio based Query Expansion}
\label{section:Multi-Portfolio}
To address the vocabulary mismatch problem in information retrieval~\cite{yao2025llm}, our process begins with query expansion using a Large Language Model (Qwen2.5-14B-Instruct~\cite{qwen2024qwen2}). Specifically, the LLM analyzes the intent of an initial user query using the query expansion prompt described in Appendix~\ref{fig:qe-prompt}, and extracts two distinct keyword sets: (i) a set of Primary Keywords and (ii) a set of Expanded Keywords. This prompt is carefully designed to elicit both literal and conceptual interpretations of the query, encouraging the model to surface domain-specific terms and semantically aligned paraphrases that might not appear explicitly in the user input. By doing so, the LLM effectively bridges the gap between the user’s linguistic expression and the terminology prevalent in the target corpus, leading to richer and more context-aware retrieval. The first set, (i) Primary Keywords, comprises five terms—each paired with corresponding synonyms—which act as semantic anchors to ensure subsequent searches remain highly faithful to the user’s intent and mitigate the risk of query drift~\cite{pan2025llm}. The second set, (ii) Expanded Keywords, includes three additional terms that may not appear in the original query but broaden the search scope and capture relevant documents expressed in different terminology, thereby enhancing both recall and lexical diversity in the retrieval stage.

These eight keywords are systematically combined with Boolean operators to form four distinct Query Portfolios (\textbf{\textit{Q}}, \textbf{\textit{S}}, \textbf{\textit{F}}, and \textbf{\textit{G}}), each encoding a complementary retrieval strategy. 

    \paragraph{\textbf{\textit{Q}} Portfolio.} \textbf{\textit{Q}}uery-expansion Portfolio maximizes recall by pairing each of the top three Primary Keywords (N=3) with their conceptually related Expanded Keywords using OR operators, and then combining them with AND operators, thereby enhancing semantic coverage and producing three separate sub-queries that reduce the risk of missing documents expressed in any alternative terminology.
    
    \paragraph{\textbf{\textit{S}} Portfolio.} \textbf{\textit{S}}ynonym-robust Portfolio enhances precision and robustness by incorporating synonyms for the top three Primary Keywords (N=3) with OR operators, applying prefix-based wildcard expansion to the Expanded Keywords, and filtering out noisy terms (e.g., “patent,” “dataset,” “photo”) with NOT operators; all components are joined with AND, yielding three sub-queries. 
    
    \paragraph{\textbf{\textit{F}} Portfolio.} \textbf{\textit{F}}ocus-balanced Portfolio seeks a balance between breadth and selectivity by grouping the top-3 Primary Keywords (N=3) as semantic anchors with OR operators and combining this group with the remaining two Primary Keywords (M=2) using AND, which generates two sub-queries. 

    \paragraph{\textbf{\textit{G}} Portfolio.} \textbf{\textit{G}}lobal-translation Portfolio addresses cross-lingual coverage by translating queries depending on the source language: Korean queries are translated into English and vice versa using the NLLB-200 model~\cite{costa2022no}. The translated Primary Keywords are grouped with OR operators, combined with the translated Expanded Keywords via AND, producing two multilingual sub-queries.

This design reflects complementary retrieval strategies—recall-oriented (\textbf{\textit{Q}}), precision-oriented (\textbf{\textit{S}}), balance-oriented (\textbf{\textit{F}}), and multilingual (\textbf{\textit{G}})—and demonstrates that diverse query formulations yield more comprehensive results than relying on a single query~\cite{chan2024rq, li2024dmqr}. The resulting sub-queries from all portfolios are then aggregated and passed to the next stage, Candidate Aggregation and Re-Ranking.

\subsubsection{Candidate Aggregation and Re-Ranking}
\label{section:ReRanking}
Once the query portfolios are constructed, they are executed in parallel efficiently against the document index to retrieve candidate documents. The top-ranked results from each query are merged into a shared pool, and duplicates are removed automatically based on unique identifiers such as CN (Control Number) or document titles. This produces a candidate set of approximately 270 unique items, ensuring both diversity and coverage. The re-ranking stage aims to maximize precision by using a Cross-Encoder~\cite{chen2024m3}. This architecture jointly processes the query and each candidate, allowing for a deep analysis of token-level interactions to produce highly accurate relevance scores. Despite its high computational cost, this critical step distills the initial set down to the top 50 most relevant documents, providing a final, high-quality context for the generation model.

\subsection{Citation-Grounded Generation}
Producing document-grounded answers with explicit citations is crucial for ensuring both informativeness and verifiability~\cite{liu2023evaluating}. Leveraging this principle, our generation module, termed Citation-Grounded Generation (CiteG), utilizes the Qwen2.5-14B-Instruct LLM~\cite{qwen2024qwen2} to generate these responses. This module employs a robust two-stage pipeline that couples LLM-driven text generation with rigorous automated citation verification. The first stage generates fluent and semantically consistent responses conditioned on retrieved documents. The second stage subsequently post-processes these outputs to append verifiable citations. This separation of generation from grounding allows CiteG to preserve response readability while enforcing strict factual accountability, maintaining coherence and traceability. For a detailed description of the prompts employed for this module, see Appendix~\ref{fig:a-prompt}.

\subsubsection{Document-Conditioned Response Generation}
The first stage produces a coherent answer strictly from the retrieved evidence by constructing a structured prompt that enforces a scientific writing style—title, introduction, main body, and conclusion—while explicitly constraining the model to use only the provided sources. Each retrieved document is reformatted into a compact context entry containing its title, abstract, and link metadata. To maintain efficiency and avoid context overflow, only the top 10 documents are included. The prompt further emphasizes three instruction-tuning principles~\cite{ouyang2022training}: grounding in the retrieved content, concise summarization, and language consistency with the user query (e.g., a Korean question yields a Korean answer). Finally, a templated answer format is applied to reduce variability across outputs and facilitate fair evaluation. The structured prompt is fed into the LLM, which generates a full-length draft answer—organized into title, introduction, body, and conclusion—that is fluent, semantically faithful to the retrieved documents, and consistent with the query language. The outcome of this stage is a raw prediction that remains coherent but does not yet contain explicit citations.

\subsubsection{Sentence-Level Citation Grounding}
The second stage of our framework enhances trustworthiness by post-processing the generated text and attaching citations. The process begins with sentence-level segmentation, where the answer is divided into minimal discourse units, typically at the sentence level. Next, during citation alignment, each sentence is evaluated against the candidate documents, and the model selects supporting sources as appropriate. A dedicated citation prompt enforces three rules: (i) factual claims must include citations; (ii) when evidence exists, supporting sources are drawn from the candidate set; and (iii) generic or transitional sentences without verifiable content may omit citations. 

Following alignment, marker assignment ensures that citation markers are incrementally attached, with multiple sentences referencing the same document are reuse the same marker to preserve readability and avoid redundancy. These markers are consolidated in a citation block, appended at the end of the answer to map each marker to its source link; if no citations are produced, this block remains empty. Finally, in the result serialization step, the output is stored in two fields: the Answer field containing the narrative text with inline markers (e.g., “... [1]”), and the Citation field listing the structured references (e.g., “[1] <link>”). This pipeline guarantees that the final output is a fluent, document-grounded narrative in which each factual statement can be traced to a verifiable source.

%% file: 04.exp.tex
\section{Experiments}
\subsection{Experimental Setup}

\paragraph{Dataset.}
We evaluate our proposed model, \algname{}, using the dataset from the \textit{ScienceON AI Challenge}~\cite{sai-challenge}. This dataset is composed of multilingual question-answer pairs focused on diverse scientific domains. It is specifically designed to test a model's ability to comprehend complex scientific queries and generate accurate, well-supported answers based on the provided corpus of scientific literature.

\paragraph{Evaluation Metrics.}
We assess performance using the official metrics of the challenge. The final score provides a holistic evaluation of both generation quality and system efficiency.
\begin{itemize}
    \item \textbf{Quality Metrics}: BLEU and METEOR scores are used to measure the fluency, grammatical accuracy, and semantic adequacy of the generated answers against the reference answers.
    \item \textbf{Retrieval Metric}: To evaluate the alignment between the cited documents and the model’s supporting evidence, we compute the average cosine similarity between each citation sentence and its corresponding document’s title–abstract pair using a SentenceTransformer model (\texttt{all-MiniLM-L6-v2})~\cite{reimers-gurevych-2019-sentence} to measure consistent semantic correlation across all generated outputs.
    \item \textbf{Efficiency Metric}: Execution Time measures the total latency from receiving a query to generating the final answer, reflecting the overall efficiency of the retrieval and generation pipeline.
\end{itemize}
For all experiments, we additionally report an overall performance score computed as follows:
\[
\text{Final Score} = \frac{1}{2} \left[ \text{mean}(\text{BLEU}, \text{METEOR}, \text{Retrieval Accuracy}) + \tilde{T}_{\text{exec}} \right],
\]
\vspace{-0.1em}
\noindent
where $\tilde{T}_{\text{exec}}$ denotes the normalized execution time scaled to [0, 1] for fair comparison across models.

\vspace{-0.1em}
\paragraph{Implementation Details.}
All experiments were conducted within the environment detailed in Table~\ref{tab:exp_environment}. We used the vLLM~\cite{kwon2023efficient} for optimized inference and memory-efficient decoding. Both the LLM and reranker operated on a single NVIDIA RTX A6000 GPU, utilizing nearly its full memory capacity.

\vspace{-0.1em}
\begin{table}[h]
\centering
\caption{Experimental environment and hardware resource utilization.}
\label{tab:exp_environment}
\setlength{\tabcolsep}{6pt}
\renewcommand{\arraystretch}{1.1}
\begin{tabular}{ll}
\toprule
\textbf{Component} & \textbf{Specification / Usage} \\
\midrule
Operating System & Ubuntu 20.04.6 LTS \\
GPU Model & NVIDIA RTX A6000 (48 GB VRAM) \\
GPU Memory Usage & 47.9 GB / 49.1 GB (97.6\%) \\
CUDA / PyTorch & 12.2 / 2.7.1 \\
LLM Model & Qwen2.5-14B-Instruct (≈42 GB usage) \\
Reranker Model & BGE-Reranker-v2-m3 (≈2 GB usage) \\
Execution Time & ≈0.7 h (50 queries) \\
\bottomrule
\end{tabular}
\end{table}

\vspace{-0.1em}
\subsection{LLM Comparison}
Table~\ref{tab:ablation_llm} compares the performance of various LLM backbones integrated into the \algname{} framework. Three representative command-tuned models with different parameter scales—Qwen-2.5~\cite{qwen2024qwen2}, Llama3.1-8B~\cite{grattafiori2024llama}, and the Korean-adapted KONI-Llama3.1-8B~\cite{KISTI-KONI}—were evaluated under identical retrieval and citation verification settings. As shown in Table~\ref{tab:ablation_llm}, Qwen-2.5-14B~\cite{qwen2024qwen2} achieved the highest overall performance across all quality metrics except efficiency.

We first examine the parameter-dependent behavior within the Qwen-2.5 series~\cite{qwen2024qwen2}. Although performance does not increase monotonically with model size, Qwen-2.5-14B~\cite{qwen2024qwen2} consistently delivers the strongest overall performance, while Qwen-2.5-3B~\cite{qwen2024qwen2} surpasses Qwen-2.5-7B~\cite{qwen2024qwen2} in both efficiency and retrieval balance. The degraded performance of Qwen-2.5-7B~\cite{qwen2024qwen2} can be attributed to tuning bias and insufficient adaptation to Korean-language data, suggesting that larger capacity alone does not guarantee better results when fine-tuning coverage is limited, especially for retrieval tasks.

Meanwhile, both Llama3.1-8B~\cite{grattafiori2024llama} and KONI-Llama3.1-8B~\cite{KISTI-KONI} demonstrated adequate citation performance under consistent evaluation settings. Specifically, KONI-Llama3.1-8B~\cite{KISTI-KONI} outperformed its base model, Llama3.1-8B~\cite{grattafiori2024llama}, on Korean-language queries. However, its overall performance remained below that of Qwen-2.5-14B~\cite{qwen2024qwen2}, which exhibited more stable citation quality and stronger multilingual robustness across diverse question domains (Appendix~\ref{comllm}). Considering this trade-off between accuracy and efficiency, we adopt Qwen-2.5-14B~\cite{qwen2024qwen2} as the backbone for subsequent experiments.

\begin{table}[h]
\centering
\caption{Overall performance comparison of different LLM backbones integrated into the \algname{} framework.}
\label{tab:ablation_llm}
\begin{tabular}{l|cccc|c}
\toprule
\textbf{Model} & \textbf{BLEU} & \textbf{METEOR} & \textbf{Retrieval Acc.} & \textbf{$\mathbf{T_{\text{exec}}}$ (s)↓} & \textbf{Final Score} \\
\midrule
Qwen-2.5-1.5B~\cite{qwen2024qwen2}  & 0.0711 & 0.2425 & 0.0113 & 784.32 & 0.4204\\
Qwen-2.5-3B~\cite{qwen2024qwen2}    & \underline{0.1643} & \underline{0.3178} & 0.3904 & \underline{793.06} & 0.4530\\
Qwen-2.5-7B~\cite{qwen2024qwen2}    & 0.0791 & 0.1762 & 0.2127 & \textbf{759.92} & 0.4009\\
Llama3.1-8B~\cite{grattafiori2024llama} & 0.0748 & 0.1881 & 0.2673 & 1562.17 & 0.4539\\
KONI-Llama3.1-8B~\cite{KISTI-KONI}  & 0.1208 & 0.2923 & \underline{0.4788} & 2605.66 & \underline{0.5044}\\
Qwen-2.5-14B~\cite{qwen2024qwen2}   & \textbf{0.2254} & \textbf{0.3959} & \textbf{0.4943} & 2582.76 & \textbf{0.5415}\\
\bottomrule
\end{tabular}
\end{table}

\subsection{Query Portfolio Comparison}
We conducted an ablation study to evaluate the contribution of each component in the proposed query portfolio mechanism.
This mechanism comprises four complementary modules: \textbf{\textit{Q}} (query-expansion), \textbf{\textit{S}} (synonym-robust), \textbf{\textit{F}} (focus-balanced), and \textbf{\textit{G}} (global-translation).
As summarized in Table~\ref{tab:ablation_qsfg}, the full configuration (\textit{\textbf{Q+S+F+G}}) achieves consistent performance across all quality metrics, demonstrating the effectiveness of integrating diverse query perspectives for robust document retrieval.

Removing any single component generally leads to a degradation in quality metrics—particularly BLEU and retrieval accuracy—indicating that each module contributes distinctively to query diversification and evidence alignment.
Among them, omitting the \textbf{\textit{Q}} block causes the most significant performance drop, underscoring the importance of query expansion in capturing semantically related contexts.
Conversely, the \textbf{\textit{S}} block has the least impact, suggesting that synonym robust offers only marginal benefits when query expansion and translation are already incorporated.
However, excluding this block reduces execution time, implying that synonym matching increases retrieval latency.

Consequently, while the configuration without \textbf{\textit{S}} block achieves a slightly higher efficiency-adjusted score, we adopt the full portfolio as the primary design, given its superior retrieval accuracy and citation quality. Overall, these results demonstrate that the query portfolio design leverages semantic, linguistic, and translational diversity to improve both retrieval precision and generation fluency.

\begin{table}[h]
\centering
\caption{Ablation study on multi-query blocks (\textbf{\textit{Q, S, F, G}}). The Δ column indicates the performance difference compared to the full model.}
\begin{tabular}{l|cccc|c}
\toprule
\textbf{Model Variant} & \textbf{BLEU }& \textbf{METEOR} & \textbf{Retrieval Acc.} & \textbf{$\mathbf{T_{\text{exec}}}$ (s)↓} & \textbf{Δ(Final Score)} \\
\midrule
Full (\textbf{\textit{Q+S+F+G}})             & \textbf{0.2254} & \underline{0.3959} & \textbf{0.4943} & 2582.76  & 0.5415\\
–\textbf{\textit{Q}} (no Query-expansion)    & 0.2080 & 0.3842 & 0.4330 & \textbf{2335.25} & -0.0729 \\
–\textbf{\textit{S}} (no Synonym-robust)     & \underline{0.2210} & 0.3895 & 0.4869 & \underline{2413.40} & -0.0002 \\
–\textbf{\textit{F}} (no Focus-balanced)     & 0.2050 & 0.3721 & 0.4490 & 2464.44 & -0.0126 \\
–\textbf{\textit{G}} (no Global-translation) & 0.2108 & \textbf{0.3965} & \underline{0.4894} & 2508.13 & -0.0330 \\
\bottomrule
\end{tabular}
\label{tab:ablation_qsfg}
\end{table}

\subsection{Case Study}
This qualitative case study complements our quantitative analysis, evaluating the MPR-CiteG model via two scenarios. The Success Case demonstrates that query-centric expansion enables high-fidelity synthesis across multiple citations while effectively avoiding hallucination. The Error Case reveals limitations with complex, multi-faceted queries. This analysis confirms the effectiveness of citation-based grounding and motivates a strategic shift from static portfolio strategies to dynamic query expansion.

\subsubsection{Correct case}

The success case in Table $\text{\ref{tab:qa_correct_example}}$ demonstrates that MPR-CiteG operates beyond simple answer retrieval, functioning as a robust mechanism to effectively control LLM hallucination and ensure factual verifiability throughout the generation process. MPR-CiteG accurately answered a query on a diabetes detection study, providing its methodology and key evaluation metrics such as 95.7\% and 95.1\%.

The strengths of the CiteG module become evident in this process. First, a cited sentence explicitly indicates the model grounded the information in a retrieved document. This suggests the LLM performed a secondary verification during the generation process, allowing the user to trace the citation and immediately determine the claim is not a hallucination. Second, sentences lacking a citation are categorized as general statements intended to guide the narrative flow. Conversely, if a specific factual claim were presented without a citation, it would provide a strong basis for suspecting hallucination.

Consequently, the CiteG module segregates every sentence in the generated response into either a `verifiable fact' (cited) or a `general statement' (uncited). This success case demonstrates that CiteG is a robust module for controlling LLM hallucinations and empowering users to assess reliability.

\begin{table}[h]
\centering
\caption{Success Case : Correct QA}
\begin{tabular}{p{1.5cm}p{13cm}} 
\toprule
\textbf{Category} & \textbf{Content} \\
\midrule

\textbf{Question} & Could you briefly summarize the methodology and results of using CNN, LSTM, and SVM combinations for noninvasive diabetes detection based on HRV signals? \\
\midrule
\textbf{MPR-CiteG Answer} &

\textbf{Title:} \newline
Using CNN, LSTM, and SVM Combinations for Noninvasive Diabetes Detection Based on HRV Signals \newline 

\textbf{Introduction:} \newline
The research focuses on the application of deep learning techniques, specifically Convolutional Neural Networks (CNN), Long Short-Term Memory (LSTM) networks, and their combinations, along with Support Vector Machines (SVM), for the noninvasive detection of diabetes using Heart Rate Variability (HRV) signals derived from ECG data. [1] The methodologies involve extracting complex temporal dynamic features from HRV data using these deep learning architectures before passing them into SVM for classification. [1] \newline

\textbf{Main Body:} \newline
The methodology involves splitting the database into training and testing sets for initial classification. [2] The CNN-LSTM combination is particularly highlighted for its ability to capture complex temporal dynamics in HRV signals. [3] The study reports a maximum accuracy of 95.7\% for the classification system using SVM after feature extraction with CNN and CNN-LSTM architectures. [3] Additionally, the CNN-LSTM model achieved an accuracy of 95.1\% using 5-fold cross-validation, showcasing its effectiveness in distinguishing between diabetic and normal HRV signals. [2]  \newline

\textbf{Conclusion:} \newline
In summary, the research demonstrates that the integration of CNN, LSTM, and SVM provides a robust framework for noninvasive diabetes detection using HRV signals. [3] The proposed method achieves high accuracy, making it a valuable tool for clinical diagnosis and early detection of diabetes. [3] The findings suggest that deep learning techniques, especially the CNN-LSTM combination, offer significant potential for improving the accuracy of diabetes detection systems. [4] \\
\midrule
\textbf{Citations} &
[1] Accurate Approach to Diabetes Detection Using Deep Learning Algorithms \newline
[2] Automated detection of diabetes using CNN and CNN-LSTM network and heart rate signals \newline
[3] Diabetes detection using deep learning algorithms \newline
[4] Hybrid CNN-LSTM for Predicting Diabetes: A Review
  \\
\bottomrule
\end{tabular}
\label{tab:qa_correct_example} 
\end{table}

\subsubsection{Incorrect case}

The Error Case (Table $\text{\ref{tab:qa_incorrect_example}}$) failed to address the query's structural requirement. It did not provide the demanded explanation of the HMM and NN integration strategy, focusing instead on functional contributions such as the HMM's probabilistic refinement and the NN's role in weight determination.

This failure stems from prioritizing frequent keywords (e.g., ``recognition rate increase''), which are heavily skewed toward a single query facet and limit contextual diversity. Furthermore, the CiteG module is constrained to generating answers based only on the Top-10 retrieved documents.

To address this limitation, the static portfolio strategy should evolve into a dynamic mechanism that interleaves reasoning and retrieval, such as IRCOT~\cite{trivedi2022interleaving}. In this framework, the model iteratively uses the output of each reasoning step as a refined query for the next retrieval, repeating the process until the original query is fully satisfied. This step-by-step strategy prevents the document pool from being skewed and ensures all query facets are included. Moreover, the CiteG module should be improved to selectively reference the most relevant documents, rather than relying on a fixed Top-10 set.

Consequently, this failure case demonstrates the limitations of the current static MPR-CiteG pipeline in handling complex, multi-faceted requirements. It highlights the need for future work to dynamically enhance the MPR stage via reasoning and expand CiteG into an adaptive generation module.

\begin{table}[h]
\centering
\caption{Error Case : Incorrect QA}
\begin{tabular}{p{1.5cm}p{13cm}} 
\toprule
\textbf{Category} & \textbf{Content} \\
\midrule
\textbf{Question} & Can you briefly explain how the hidden Marov model and the neural network integration strategy were organized to increase the recognition rate of audiovisual voice recognition in the noise environment? \\
\midrule
\textbf{MPR-CiteG Answer} &
\textbf{Title:}  \newline
Enhanced Audiovisual Speech Recognition Accuracy using a Hidden Markov Model and Neural Network Integration Strategy
\newline 
\textbf{Introduction:} \newline
This study proposes a strategy that integrates Hidden Markov Models (HMMs) and Neural Networks (NNs) to enhance the performance of audiovisual speech recognition in noisy environments. [1] This method utilizes both visual and auditory information for speech recognition, demonstrating superior performance, especially in highly noisy conditions. [1]   \newline
\textbf{Main Body:} \newline
The study improved the recognition performance using visual information by refining the probabilistic optimization algorithm of the Hidden Markov Model. [1] Furthermore, speech modeling was enhanced by considering the dynamic characteristics of speech, which improves robustness in noisy environments. [1] Finally, a neural network was utilized in the information integration phase to maximize the complementarity between visual and auditory information. [1] This process yields optimal recognition results by outputting appropriate weights for the given audiovisual data. [1] \newline
\textbf{Conclusion:}  \newline
This research proposes a strategy integrating a Hidden Markov Model and a Neural Network to improve the performance of speech recognition systems. [1] This method enhances robustness in noisy environments and maximizes the complementarity between visual and auditory information, thereby improving the final recognition rate. [1] Experimental results showed that the proposed system achieved higher performance than existing systems in various noisy environments. [1] \\
\midrule
\textbf{Citations} &
[1] Audio-Visual Speech Recognition : Stochastic Optimization of HIdden Markov Models, Modeling of Interframe Correlations, and Integration with Neural Networks \\
\bottomrule
\end{tabular}
\label{tab:qa_incorrect_example} 
\end{table}

%% file: 05.con.tex
\section{Conclusion and Future Work}
In this work, we presented \algname{}, a framework integrating Multi-Portfolio Retriever (MPR) with Citation-Grounded Generation (CiteG), which ranked second place in the ScienceON AI Challenge. Comprehensive experiments and ablation studies validated our approach, confirming that the synergistic design of MPR—combining query expansion, synonym robust, focus-balanced, and global translation—significantly improves retrieval accuracy and contextual coverage.

However, our failure analysis revealed that the current static pipeline struggles with complex, multi-faceted queries by retrieving biased documents, thereby limiting generation to a fixed Top-10 evidence pool. Future work should evolve \algname{} into a dynamic mechanism. We propose a stepwise retrieval strategy that interleaves reasoning with retrieval to prevent bias, while also developing CiteG into an adaptive generation module that selectively references the most relevant documents. This will be crucial for building a robust, scalable RAG system capable of handling complex information needs.

%% file: 06.app.tex
\section{Appendix}

\tcbset{
  colback=gray!5,      
  colframe=black!80,   
  coltitle=white,      
  fonttitle=\bfseries, 
  boxrule=1.5pt,       
  arc=5pt,             
  top=5pt,bottom=5pt,  
  left=7pt,right=7pt,
  toptitle=2pt,         
  bottomtitle=2pt,      
  fonttitle=\bfseries,
  fontupper=\small,
}

\subsection{MPR Prompts}
\noindent
The \textbf{Query Expansion Prompt} instructs the model to extract and expand key search terms from a scientific question, ensuring that the generated queries remain faithful to the original intent while being enriched with relevant scientific concepts. The instruction template is presented as follows:

\begin{tcolorbox}[title=Query Expansion Prompt, coltitle=white, colback=gray!5, colframe=black!70, label={fig:qe-prompt}]
\textbf{Role:} You are a research assistant who designs search keywords for scientific questions.  
Follow the rules below strictly to generate keywords and search queries.  

\textbf{Rules:}
\begin{enumerate}
  \item \textbf{Primary keywords (5):} Must be terms that appear in the question or very close synonyms.  \newline
  \hspace{1.9em}– Rank them in order of importance (most important first).  \newline
  \hspace{1.9em}– Maintain scientific accuracy, use noun phrases, and avoid duplicates.
  \item \textbf{Expanded keywords (3):} May not appear in the question but must be relevant to the intent, and broaden the scope with higher-level concepts, related phenomena, standard terms, or synonyms.
  \item \textbf{Output must be JSON only.} Do not add any explanations.
\end{enumerate}
\end{tcolorbox}

\subsection{CiteG Prompts}
\noindent
The \textbf{Answer Prompt} guides the model to generate concise and fact-grounded answers based solely on the retrieved documents, maintaining the same language as the user's question for clarity and coherence. The instruction template is presented as follows:

\begin{tcolorbox}[title=Answer Prompt, coltitle=white, colback=gray!5, colframe=black!70, label={fig:a-prompt}]
\textbf{Role:} You are a meticulous Question \& Answering chatbot in the scientific domain. Your task is to answer the user's question strictly based on the provided documents.

\textbf{Rules:}
\begin{enumerate}
  \item Read the user's question and the \textbf{retrieved documents} carefully.
  \item Formulate a clear and concise answer using the information \textbf{present in the documents.}
  \item Language policy: in the \textbf{SAME language} as the user's question (English → English, Korean → Korean). Generate the entire output.
\end{enumerate}
\end{tcolorbox}

\noindent
The \textbf{Citation Prompt} enforces factual grounding by instructing the model to identify the most relevant supporting document for each sentence, thereby ensuring clear source-level verifiability and consistency. The instruction template is presented as follows:

\begin{tcolorbox}[title=Citation Prompt, coltitle=white, colback=gray!5, colframe=black!70, label={fig:c-prompt}]
\textbf{Role:} You are a meticulous assistant that verifies if a given sentence is directly supported by a list of source documents. Your task is to find the SINGLE source document that provides a DIRECT and VERIFIABLE origin for the sentence.

\textbf{Rules:}
\begin{enumerate}
  \item Prioritize finding a \textbf{citation.} Your main goal is to \textbf{ground the text.} You should cite any sentence that contains specific information (names, numbers, key concepts, factual claims) that can be reasonably traced back to a single source document.
  \item A citation MUST be added if the sentence contains a key \textbf{factual} claim. Do not be overly strict; if the information plausibly comes from a source, cite it.
  \item Only respond with 0 for sentences that are purely generic, introductory, or transitional and lack specific, verifiable information from the sources.
\end{enumerate}
\end{tcolorbox}

\subsection{Comparative Model Performance Analysis }
\label{comllm}

Tables \ref{tab:qa_koni_example} and \ref{tab:qa_qwen_example} compare the performance of KONI-Llama3.1-8B~\cite{KISTI-KONI} and the Qwen-2.5 series~\cite{qwen2024qwen2} on Korean queries. Llama3.1-8B~\cite{grattafiori2024llama} and Qwen-2.5-7B were excluded from the comparison, as they either failed to perform retrieval for most Korean queries or generated responses in Chinese.

As shown in Table \ref{tab:qa_koni_example}, KONI-Llama3.1-8B demonstrates outstanding performance in Korean, accurately capturing the key reinforcing factors and underlying research findings. Furthermore, Table \ref{tab:qa_qwen_example} illustrates that as the number of parameters in the Qwen-2.5 series increases, its outputs become more stable, providing well-structured responses and sufficient citations.

This study imposed no specific parameter limit. Therefore, despite KONI-Llama3.1-8B's excellent Korean generation capabilities, the Qwen-2.5-14B was selected as the final backbone after comprehensively considering its multilingual generation capabilities and overall citation quality. Looking ahead, if cost-effectiveness becomes a significant constraint, leveraging KONI-Llama3.1-8B's strong Korean performance to build a lightweight model would represent a promising direction.

\begin{table}[htp!]
\centering
\caption{Example of QA: KONI}
\label{tab:qa_koni_example}
\begin{tabular}{p{1.5cm}p{13cm}}
\toprule
\textbf{Category} & \textbf{Content} \\
\midrule
\textbf{Question} & 한국 IT융합 신산업 역량강화요인과 이를 도출하기 위한 연구 결과를 간결하게 요약해 주시겠습니까? \\
\midrule
\textbf{KONI \newline Answer} &
\textbf{Title:} 한국 IT융합 신산업 역량강화요인 \newline
\textbf{Introduction:} 본 연구는 한국의 IT융합 신산업 역량강화요인을 도출하기 위한 정책연구를 수행하였다. [1] 2011년도에 실시한 전문가 델파이 조사 결과를 분석하여 한국의 IT융합역량의 수준 변화를 평가하였다. 이를 통해 한국의 IT융합 신산업 역량강화요인을 도출하였다. [1] \newline
\textbf{Main Body:} 전문가 델파이 조사 결과, 한국의 IT융합 신산업 역량강화요인으로는 다음과 같은 요소들이 도출되었다. 첫째, 기술개발과 혁신 역량 강화가 필요하다. 둘째, 인력 양성과 인재 확보가 중요하다. [2] 셋째, 산업체와 학계, 정부 간의 협력이 필요하다. 넷째, 투자 확대와 자금 조달이 필요하다. [1] 다섯째, 법적·제도적 지원이 필요하다. 여섯째, 시장 확대와 수출 활성화가 필요하다. [3] 이러한 요소들은 한국의 IT융합 신산업 역량강화에 중요한 역할을 할 것으로 판단된다. [1] \newline
\textbf{Conclusion:} 본 연구는 한국의 IT융합 신산업 역량강화요인을 도출하기 위한 정책연구를 수행하였다. [1] 전문가 델파이 조사 결과, 기술개발과 혁신 역량 강화, 인력 양성과 인재 확보, 산업체와 학계, 정부 간의 협력, 투자 확대와 자금 조달, 법적·제도적 지원, 시장 확대와 수출 활성화 등이 한국의 IT융합 신산업 역량강화요인으로 도출되었다. [1] 이러한 요소들은 한국의 IT융합 신산업 역량강화를 위한 중요한 방향으로 판단된다. [1] \\
\bottomrule
\end{tabular}
\end{table}

\clearpage

\begin{table}[htp!]
\centering
\caption{Example of QA: Qwen 1.5B, 3B and 14B}
\label{tab:qa_qwen_example}
\begin{tabular}{p{1.5cm}p{13cm}}
\toprule
\textbf{Category} & \textbf{Content} \\
\midrule
\textbf{Question} & 한국 IT융합 신산업 역량강화요인과 이를 도출하기 위한 연구 결과를 간결하게 요약해 주시겠습니까? \\
\midrule
\textbf{Qwen 1.5B \newline Answer} &
IT융합 신산업 역량강화요인은 다음과 같습니다:

\textbf{1. 국내 IT 산업의 역량}: 국내 IT 산업은 기술경쟁력을 바탕으로 산업자체를 고도화하고 다른 산업의 생산성과 효율성을 강화하기 위해 융합과 복합을 촉진하고 차세대 기술과의 결합을 통한 신산업을 창출해야 하는 막대한 과제가 주어져 있습니다. \textbf{2. 국내 IT 산업의 역량분석}: 국내 IT 산업의 역량은 다른 국가에 비해 상대적으로 우위에 있지만 부문별로 그 차이가 나타나고 있습니다. 특히 글로벌 유수평가 기관의 국내 IT 산업에 대한 평가보고서는 매년 그 변동이 심하여 일관성이 부족합니다. \textbf{3. IT융합 신산업육성 및 선도방안}: 정부와 기업을 중심으로 IT융합을 촉진하기 위한 정책방향과 전략적 시사점을 제시하였습니다. 이를 위해 IT융합을 촉진하기 위한 산업 생태계 마련을 위해 정부의 규제 완화와 벤처 육성정책 확대, 키스톤 기업의 협업기반 융합과제 확대와 경험 공유, 벤처기업 등 기술개발 주체의 기술역량 강화, 그리고 크로스오버형 인재양성을 위한 대학의 역할이 필요하다고 합니다. \textbf{4. 중소기업 IT 융합 역량강화 방안}: 중소기업의 경우 융합에 대한 관심은 높으나 구체적인 추진은 부족하여, 이에 대한 정책적인 고려가 요구됩니다. 창조적인 아이디어가 가미된 융합은 중소기업에게 성공을 위한 새로운 기회를 제공할 수 있으므로 대기업에 비해 상대적으로 자원이 부족한 중소기업도 융합 아이디어만 있으면 부가가치를 높이고 신사업을 창출함으로써 성공 가능하다고 합니다. \\
\midrule
\textbf{Qwen \newline 3B \newline Answer} &
\textbf{Title:} 한국 IT융합 신산업 역량강화요인과 이를 도출하기 위한 연구 결과 \newline
\textbf{Introduction:} 이 논문은 IT융합 신산업의 역량강화요인을 도출하기 위해 2011년에 실시된 정책연구의 학계 및 공공기관 전문가 델파이 결과를 분석하고, 이를 선행연구들과 비교하여 한국의 IT 융합역량의 수준 변화를 평가하였다. \newline
\textbf{Main Body:} 논문은 IT융합 신산업의 역량강화요인을 도출하기 위해 2011년에 실시된 정책연구의 델파이 결과를 분석하였으며, 이를 선행연구들과 비교하여 한국의 IT 융합역량의 수준 변화를 평가하였다. [1] 이 연구는 IT융합 신산업의 역량강화요인을 도출하고자 하였으며, 이를 통해 한국의 IT융합 신산업의 발전을 위한 정책 제언을 제공하였다. [1] \newline
\textbf{Conclusion:} 이 논문은 IT융합 신산업의 역량강화요인을 도출하고, 이를 통해 한국의 IT융합 신산업의 발전을 위한 정책 제언을 제공하였다. [1] 이는 한국의 IT융합 신산업의 역량강화를 위한 중요한 기초 자료가 되었다. \\
\midrule
\textbf{Qwen 14B \newline Answer} &
\textbf{Title:} 한국 IT융합 신산업 역량강화요인과 이를 도출하기 위한 연구 결과 요약 \newline
\textbf{Introduction:} 이 연구는 한국의 IT융합 신산업 역량강화요인을 도출하기 위해 2011년 정책연구의 학계 및 공공기관 전문가 델파이 결과를 분석하였다. [1] IT융합은 IT의 고도화를 바탕으로 다른 산업의 발전을 견인하거나 새로운 산업을 창출하는 것으로, 한국을 포함한 주요국가들이 이를 중요하게 여긴다. [2] \newline
\textbf{Main Body:} 연구는 IT융합 신산업 역량강화요인을 도출하기 위해 정부와 기업의 역할, 정책방향, 전략적 시사점을 제시하였다. [3] 이를 위해 정부의 규제 완화, 벤처 육성정책 확대, 키스톤 기업의 협업기반 융합과제 확대, 크로스오버형 인재양성 등이 필요하다는 결론을 내렸다. [4] 또한, IT융합 지원시스템 구축을 통해 IT기업과 비IT기업 간의 협업을 촉진하고, 기업정보 전산시스템(DB)을 구축하여 IT융합사업 참여를 돕는 지원시스템을 마련해야 함을 제안하였다. [5] \newline
\textbf{Conclusion:} IT융합 신산업 역량강화를 위해서는 정부와 기업 모두의 역할이 중요하며, 이를 위해 정부는 규제 완화와 벤처 육성정책을 강화하고, 기업은 협업을 통해 새로운 기술과 산업을 창출해야 한다. [4] 또한, IT융합 지원시스템 구축을 통해 기업 간 협업을 촉진하고, IT융합사업 참여를 돕는 인프라를 마련해야 한다. [5] 이러한 노력은 한국의 IT융합 신산업 역량을 강화하고, 경쟁력을 향상시키는 데 기여할 것이다. [2] \\
\bottomrule
\end{tabular}
\end{table}